\documentclass{article} 
\usepackage{nips13submit_e,times}
\usepackage{hyperref}
\usepackage{url}

\usepackage{graphicx}
\usepackage{caption}
\usepackage{subcaption}
\usepackage{amsmath}
\usepackage{amssymb}
\usepackage{enumitem}
\usepackage{algorithm}
\usepackage{multirow}

\usepackage{natbib}

\renewcommand{\vec}[1]{\mathbf{#1}}
\newcommand{\NAcell}{\multicolumn{2}{c|}{\scriptsize$\mbox{N/A}$}}
\title{Twitter-Network Topic Model: A Full Bayesian Treatment for Social Network and Text Modeling}

\author{
Kar Wai Lim \\
ANU, NICTA \\
Canberra, Australia \\
\And
Changyou Chen \\
ANU, NICTA \\
Canberra, Australia \\
\And
Wray Buntine \\
NICTA, ANU \\
Canberra, Australia \\
}

\nipsfinalcopy 

\begin{document}

\maketitle

\begin{abstract}
Twitter data is extremely noisy -- each tweet is short, unstructured and with informal 
language, a challenge for current topic modeling. On the other hand, tweets are 
accompanied by extra information such as authorship, hashtags and the user-follower 
network. Exploiting this additional information, we propose the {\em Twitter-Network} (TN) 
topic model to jointly model the text and the social network in a full Bayesian 
nonparametric way. The TN topic model employs the hierarchical Poisson-Dirichlet processes 
(PDP) for text modeling and a Gaussian process random function model for social network 
modeling. We show that the TN topic model significantly outperforms several existing 
nonparametric models due to its flexibility.
Moreover, the TN topic model enables additional informative inference such as authors' 
interests, hashtag analysis, as well as leading to further applications such as 
author recommendation, automatic topic labeling and hashtag suggestion. Note our 
general inference framework can readily be applied to other topic models with embedded 
PDP nodes. 

%
\end{abstract}

\section{Introduction}
\label{sec:introduction}

Emergence of web services such as blog, microblog and social networking websites allows 
people to contribute information publicly. This user-generated information is 
generally more personal, informal and often contains personal opinions. In 
aggregate, it can be useful for reputation analysis of entities and products, natural 
disasters detection, obtaining first-hand news, or even demographic analysis. Twitter, an 
easily accessible source of information, allows users to voice their opinions and thoughts 
in short text known as {\emph{tweets}}. 

Latent Dirichlet allocation (LDA)~\citep{blei2003latent} is a popular form of topic model. 
Unfortunately, a direct application of LDA on tweets yields poor result as tweets are 
short and often noisy~\citep{Zhao:2011:CTT:1996889.1996934}, {\it{i.e.}}\ tweets are 
unstructured and often contain grammatical and spelling errors, as well as 
{\emph{informal}} words such as user-defined abbreviations due to the 140 characters 
limit. LDA fails on short tweets since it is heavily dependent on word co-occurrence. Also 
notable is that text in tweets may contain special tokens known as {\em{hashtags}}; they 
are used as keywords and allow users to link their tweets with other tweets tagged with 
the same hashtag. Nevertheless, hashtags are informal since they have no standards. Hashtags can be 
used as both inline words or categorical labels. Hence instead of being hard labels, hashtags are 
best treated as special words which can be the themes of the tweets. Tweets 
are thus challenging for topic models, and {\it ad hoc} alternatives are used instead.
In other text analysis applications, tweets are often `cleansed' by NLP methods such as lexical normalization~\citep{baldwinnoisy}. However, the use of normalization is also criticized~\citep{Eisenstein2013}.

In this paper, we propose a novel method for short text modeling by leveraging 
the auxiliary information that accompanies tweets. This information, complementing word 
co-occurrence, allows us to model the tweets better, as well as opening the door to more 
applications, such as user recommendation and hashtag suggestion. Our main contributions 
include:
1)~a fully Bayesian nonparametric model called {\em Twitter-Network} (TN) 
{\em topic model} that models tweets very well; and
2)~a combination of both the {\em hierarchical Poisson Dirichlet process} (HPDP) and the 
{\em Gaussian process} (GP) to jointly model text, hashtags, authors and the followers 
network.
We also develop a flexible framework for arbitrary PDP networks, which allows quick 
deployment (including inference) of new variants of HPDP topic models.
Despite the complexity of the TN topic model, its implementation is made relatively 
straightforward with the use of the framework.

\vspace{-1mm}
\section{Background and Related Work}
\label{sec:background}


LDA is often extended for different types of data, some notable 
examples that use auxiliary information are 
the {\em author-topic model} \citep{rosen2004author}, 
the {\em tag-topic model} \citep{tsai2011tag}, 
and 
{\em Topic-Link LDA} \citep{liu2009topic}. 
However, these models only deal with just one kind of additional information and do not 
work well with tweets since they are designed for other types of text data. Note that the 
tag-topic model treats tags as hard labels and uses them to group text documents, which is 
not appropriate for tweets due to the noisy nature of hashtags. 
{\em{Twitter-LDA}} \citep{Zhao:2011:CTT:1996889.1996934} and 
the {\em behavior-topic model} \citep{qiu2013not} were designed to explicitly model 
tweets. Both models are not admixture models since they limit one 
topic per document. The behavior-topic model analyzes tweets' ``posting 
behavior'' of each topic for user recommendation. On the other hand, 
the {\em biterm topic model} \citep{Yan:2013:BTM:2488388.2488514} uses only the biterm 
co-occurrence to model tweets, discarding document level information. Both biterm topic 
model and Twitter-LDA do not incorporate any auxiliary information. All the above topic 
models also have a limitation in that the number of topics need to be chosen in advance, 
which is difficult since this number is not known.

To sidestep the need of choosing the number of topics, \citep{TehJor2010a} proposed 
{\em Hierarchical Dirichlet process} (HDP) LDA, which utilizes the Dirichlet process (DP) 
as nonparametric prior. Furthermore, one can replace the DP with the Poisson-Dirichlet 
process (PDP, also known as the Pitman-Yor process), which models the power-law of word 
frequencies distributions in natural languages. In natural languages, the distribution of 
word frequencies exhibits a power-law \citep{goldwater2006interpolating}. 
For topic models, replacing the Dirichlet distribution with the PDP can yield great 
improvement \citep{Sato:2010:TMP:1835804.1835890}.

Some recent work models text data with network information 
(\citep{liu2009topic,chang2010hierarchical,NallapatiAX:KDD08}),
however, these models are parametric in nature and can be restrictive.
On the contrary, Miller {\it{et al.}}\,\citep{miller2009nonparametric} and Lloyd {\it{et al.}}\,\citep{lloyd2012random} model network data directly with nonparametric priors, {\it{i.e.}}\,with the Indian Buffet process and the Gaussian process respectively, but do not model text.

\begin{figure*}[b]
\parbox{.54\linewidth}{
\begin{subfigure}{\linewidth}
  \includegraphics[width=\linewidth]{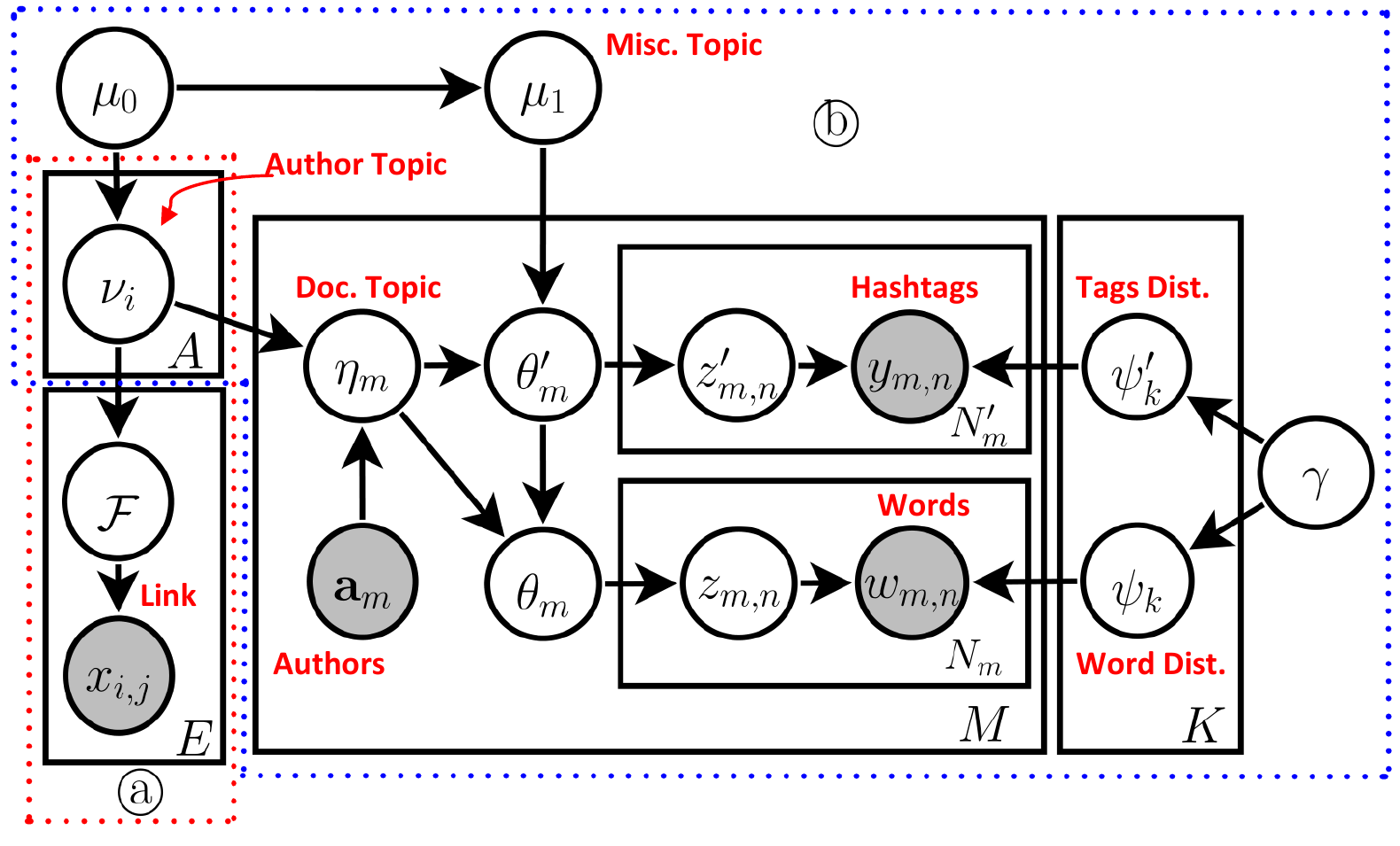}
\end{subfigure}%
	\caption{Twitter-Network topic model}
	\label{fig:author_tag}
}
\parbox{.45\linewidth}{
\begin{subfigure}{\linewidth}
	\vspace{-2.5mm}
  \includegraphics[width=\linewidth]{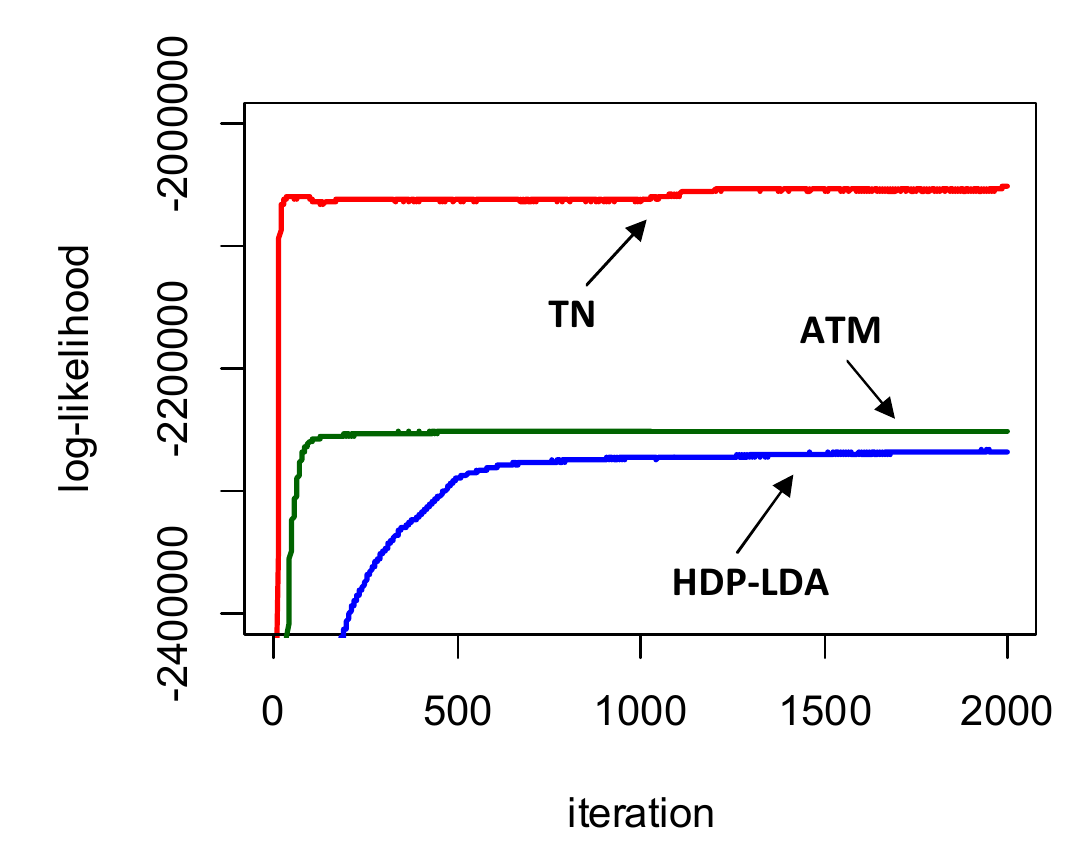}
\end{subfigure}%
\vspace{-2mm}
\caption{Log-likelihood vs. iterations}
\label{fig:train_lik}
}
\end{figure*}

\vspace{-1mm}
\section{Model Summary}
\label{sec:model}

The TN topic model makes use of the accompanying {\em hashtags}, {\em authors}, and 
{\em followers network} to model tweets better. The TN topic model is composed of two main 
components: a HPDP topic model for the text and hashtags, and a GP based random function 
model for the followers network. The authorship information serves to connect the two 
together.

We design our HPDP topic model for text as follows. First, generate the global topic 
distribution $\mu_0$ that serves as a prior. Then generate the respective authors' topic 
distributions $\nu$ for each author, and a miscellaneous topic distribution $\mu_1$ to 
capture topics that deviate from the authors' usual topics. Given $\nu$ and $\mu_1$, 
we generate the topic distributions for the documents, and words ($\eta$, 
$\theta'$, $\theta$). We also explicitly model the influence of hashtags to words.
Hashtag and word generation follows standard LDA and is not discussed here. Note that the 
tokens of hashtags are shared with the words, {\it{i.e.}}\ the hashtag {\em{\#happy}} share 
the same token as the word {\em{happy}}. Also note that all distributions on probability 
vectors are modeled by the PDP, making the model a network of PDP nodes.

The network modeling is connected to the HPDP topic model {\it{via}} the author topic 
distributions $\nu$, where we treat $\nu$ as inputs to the GP in the network model. The 
GP, denoted as $\mathcal{F}$, determines the links between the authors ($x$). 
Figure~\ref{fig:author_tag} displays the graphical model of TN, where 
region~\textcircled{a} and \textcircled{b} shows~the network model and topic model 
respectively. See supplementary material\footnote{Supplementary material is available online at the authors' websites.} for a detailed description. 
We emphasize that our treatment of the network model is different to that of 
\citep{lloyd2012random}. We define a new kernel function based on the cosine similarity in 
our network model, which provides significant improvement over the original kernel 
function. Also, we derive a new sampling procedure for inference due to the additive 
coupling of topic distributions and network connections.

\vspace{-1mm}
\section{Posterior Inference}
\label{inference}

We alternatively perform Markov chain Monte Carlo (MCMC) sampling on the topic model and 
the network model, conditioned on each other. We derive a collapsed Gibbs sampler for the 
topic model, and a Metropolis-Hastings (MH) algorithm for the network model. 
We develop a framework to perform collapse Gibbs sampling generally on any Bayesian 
network of PDPs, built upon the work of~\citep{buntine2010bayesiannetwork,
chen2011sampling}, which allows quick prototyping and development of new variants of topic 
model. We refer the readers to the supplementary materials for the technical details.

\section{Experiments and Applications}
\label{sec:experiment}

We evaluate the TN topic model quantitatively with standard topic model measures such as 
test-set perplexity, likelihood convergence and clustering measures. Qualitatively, we 
evaluate the model by visualizing the topic summaries, authors' topic distributions and 
by performing an automatic labeling task. We compare our model with HDP-LDA, a nonparametric variant of the 
author-topic model (ATM), and the original random function network model. We also perform 
ablation studies to show the importance of each component in the model. The 
results of the comparison and ablation studies are shown in Table~\ref{tbl:perplexity}.
We use two tweets corpus for experiments, first is a subset of Twitter7 
dataset\footnote{http://snap.stanford.edu/data/twitter7.html}~\citep{yang2011patterns}, 
obtained by querying with certain keywords ({\it{e.g.}}\ finance, sports, politics). we remove tweets that are not English with {\em langid.py} \citep{lui2012langid} and 
filter authors who do not have network information and who authored less than 100 tweets. 
The corpus consists of 60370 tweets by 94 authors. We then randomly select 90\% of the 
dataset as training documents and use the rest for testing. Second tweets corpus is obtained from \citep{nicta_6273}, which contains a total of 781186 tweets. We note that we perform no word normalization to prevent any loss of meaning of the noisy text.

\vspace{-2mm}
\paragraph{Experiment Settings}

In all cases, we vary $\alpha$ from $0.3$ to $0.7$ on topic nodes 
($\vec{\mu}_0$, $\vec{\mu}_1$, $\vec{\nu}_i$, $\vec{\eta}_m$, $\vec{\theta}'_m$, $\vec{\theta}_m$)
and set $\alpha = 0.7$ on vocabulary nodes ($\vec{\psi}$, $\vec{\gamma}$) to induce 
power-law. We initialize $\beta$ to $0.5$, and set its hyperprior to 
$\mbox{Gamma}(0.1, 0.1)$.
We fix the hyperparameters $\lambda$'s, $s$, $l$ and $\sigma$ to $1$ since their values have 
no significant impact on model performance. In the following evaluations, we run the 
sampling algorithms for 2000 iterations for the training likelihood to converge. We repeat 
each experiment five times to reduce the estimation error of the evaluation measures.
In the experiments for the TN topic model, we achieve a better computational efficiency by 
first running the collapsed Gibbs sampling for 1000 iterations 
before the full inference procedure. In Figure~\ref{fig:train_lik}, we can see that the TN topic model converges quickly compared to the HDP-LDA and the nonparametric ATM. Also, the training likelihood of the TN topic model becomes better sampling for the network information after 1000 iterations.

\begin{table*}[t!]
  \centering
  \parbox{.53\linewidth}{
  	\centering
    \caption{Perplexity \& network log-likelihood}
    \label{tbl:perplexity}
	  \begin{tabular}{|l|r@{\tiny$\pm$}l|r@{\tiny$\pm$}l|}
		  \hline
	  	  & \multicolumn{2}{c|}{Perplexity} & \multicolumn{2}{c|}{Network} \\ 
	  	  \hline
	  	  HDP-LDA & $ 358.1 $ & {\tiny $6.7$} & \NAcell \\ 
	  	  \hline
	  	  ATM & $ 302.9 $ & {\tiny $8.1$} & \NAcell \\ 
	  	  \hline
	  	  Random Function & \NAcell & $ -294.6 $ & {\tiny $5.9$} \\ 
	  	  \hline \hline
	  	  No Author & $ 243.8 $ & {\tiny $3.4$} & \NAcell \\ 
	  	  \hline
	  	  No Hashtag & $ 307.5 $ & {\tiny $8.3$} & $ -269.2 $ & {\tiny $9.5$} \\ 
	  	  \hline
	  	  No $\mu_1$ node & $ 221.3 $ & {\tiny $3.9$} & $ -271.2 $ & {\tiny $5.2$} \\ 
	  	  \hline
	  	  No Word-tag link & $ 217.6 $ & {\tiny $6.3$} & $ -275.0 $ & {\tiny $10.1$} \\ 
	  	  \hline
	  	  No Power-law & $ 222.5 $ & {\tiny $3.1$} & $ -280.8 $  & {\tiny $15.4$} \\ 
	  	  \hline
	  	  No Network & $ 218.4 $ & {\tiny $4.0$} & \NAcell \\ 
	  	  \hline \hline
	  	  Full TN & $ {\bf 208.4} $ & {\tiny $3.2$} & $ {\bf -266.0}$ & {\tiny $6.9$} \\ 
	  	  \hline
	  \end{tabular}
  }
  \parbox{.45\linewidth}{
  	\centering
  	\renewcommand{\arraystretch}{1.1} 
    \caption{Labeling topics with hashtags}
 	\label{tbl:exploration}
	\begin{tabular}{|l|c|}
	  	\hline
 		 & Top hashtags/words \\
	  	\hline
	  	\multirow{3}{*}{T0} & {\textbf{\#finance \#money \#economy}} \\
	  	\cline{2-2}
 		 & finance money bank marketwatch \\
 		 & stocks china group \\
	  	\hline
	  	\multirow{3}{*}{T1} & {\textbf{\#politics \#iranelection \#tcot}} \\
	  	\cline{2-2}
 		 & politics iran iranelection tcot \\
 		 & tlot topprog obama \\
	  	\hline
	  	\multirow{3}{*}{T2} & {\textbf{\#music \#folk \#pop}} \\
	  	\cline{2-2}
	  	& music folk monster head pop \\ 
	  	& free indie album gratuit \\ 
	  	\hline
	\end{tabular}
  }
\end{table*}

\vspace{-2mm}
\paragraph{Automatic Topic Labeling}\label{sec:topicexplore}

There have been recent attempts to label topics automatically in topic modeling. Here, we show that using hashtag information allows us to get good labels for topics. Table~\ref{tbl:exploration} shows topics labeled by the TN topic model. More detailed 
topic summaries are shown in the supplementary material. We empirically evaluate the 
suitability of hashtags in representing the topics and found that, consistently, 
over 90\% of the hashtags are good candidates for the topic labels. 

\vspace{-2mm}
\paragraph{Inference on Authors' Topic Distributions}

In addition to inference on the topic distribution of each document, the TN topic model 
allows us to analyze the topic distribution of each author. 
Table~\ref{tbl:authordistribution} presents a summary of topics by different authors, where topics are obvious from the Twitter ID.

\vspace{-2mm}
\paragraph{Author Recommendation}
\label{sec:linkpred}

We illustrate the use of the TN topic model for author recommendation. On a new test 
dataset with 90451 tweets and 625 new authors, we predict the most similar and dissimilar 
authors for the new authors, based on the training model of 60370 tweets. 
We quantify the recommendation quality with the cosine similarities of the authors' 
topic distributions for the recommended author pairs.
We compare our new kernel function with the original kernel 
function (denoted as {\em original}) used in~\citep{lloyd2012random}.
Table~\ref{tbl:linkpred} shows average cosine similarities between the recommended and 
not-recommended authors. This suggests that our kernel function is more appropriate.
Additionally, we manually checked the recommended authors and we found that they 
usually belong to the same community, {\it i.e.}, having tweets with similar topics.

\begin{table*}[t!]
\centering
\parbox{.57\linewidth}{
	\caption{Topics by authors}
	\label{tbl:authordistribution}
	\renewcommand{\arraystretch}{1.1} 
	\begin{tabular}{|l|l|}
	\hline
	Twitter ID & Top topics represented by hashtags \\
	\hline
	finance\_yard & \#finance \, \#money \, \#realestate \\
	\hline
	ultimate\_music & \#music \, \#ultimatemusiclist \, \#mp3 \\
	\hline
	seriouslytech & \#technology \, \#web \, \#tech \\
	\hline
	seriouspolitics & \#politics \, \#postrank \, \#news \\
	\hline
	pr\_science & \#science \, \#news \, \#postrank \\
	\hline
	\end{tabular}
}
\parbox{.40\linewidth}{
	\caption{Cosine similarity}
	\label{tbl:linkpred}
	\renewcommand{\arraystretch}{1.1} 
	\begin{tabular}{|c|c|c|c|}
	\hline
	{\scriptsize Recommended} & 1st & 2nd & 3rd \\
	\hline
	Original & $0.00$ & $0.05$ & $0.06$ \\
	\hline
	TN & {$\bf 0.78$} & {$\bf 0.57$} & {$\bf 0.55$} \\
	\hline
	{\scriptsize Not-recommended} & 1st & 2nd & 3rd \\
	\hline
	Original & $0.36$ & $0.33$ & $0.14$ \\
	\hline
	TN & {$\bf 0.17$} & {$\bf 0.09$} & {$\bf 0.10$} \\
	\hline
	\end{tabular}
}
\end{table*}

\vspace{-2mm}
\paragraph{Clustering and Topic Coherence}

We also evaluate the TN topic model against state-of-the-art LDA-based clustering 
techniques~\citep{nicta_6273}. We find that the TN topic model outperforms the state-of-the-art in 
{\em{purity}}, normalized mutual information and pointwise mutual information (PMI). Due 
to space, the evaluation result is provided in the supplementary material.

\vspace{-1mm}
\section{Conclusion and Future Work}
\label{conclusion}

We propose a full Bayesian nonparametric {\em Twitter-Network} (TN) topic model that 
jointly models tweets and the associated social network information. Our model employs a 
nonparametric Bayesian approach by using the PDP and GP, and achieves flexible modeling by 
performing inference on a network of PDPs. Our experiments with Twitter dataset show that 
the TN topic model achieves significant improvement compared to existing baselines. 
Furthermore, our ablation study demonstrates the usefulness of each component of the TN 
model. Our model also shows interesting applications such as {\em author recommendation}, 
as well as providing additional informative inferences. 

We also engineered a framework for rapid topic model development, which is important due to the complexity of the model. While we could have used 
Adaptor Grammars \citep{johnson2007adaptor}, our framework yields more efficient 
computation for topic models.

Future work includes speeding up the posterior inference algorithm, especially for the 
network model, as well as incorporating other auxiliary information that is available in 
social media such as {\em location}, {\em hyperlinks} and {\em multimedia contents}. We 
also intend to explore other applications that can be addressed with the TN topic model, 
such as {\em{hashtag recommendation}}. It is also interesting to apply the TN topic 
model to other types of data such as blog and publication data.

\section*{Acknowledgement}
We would like to thank the anonymous reviewers for their helpful feedback and comments.

NICTA is funded by the Australian Government through the Department of
Communications and the Australian Research Council through the ICT
Centre of Excellence Program.

\small
\bibliographystyle{apa}
\bibliography{citations_full,yeewhyeteh}

\end{document}